\documentclass[runningheads]{llncs}
\usepackage{graphicx}

\usepackage{tikz}
\usepackage{comment}
\usepackage{amsmath,amssymb} 
\usepackage{color}
\usepackage[colorlinks,
            linkcolor=red,
            anchorcolor=blue,
            citecolor=green
            ]{hyperref}
\usepackage[accsupp]{axessibility}  

\usepackage[width=122mm,left=12mm,paperwidth=146mm,height=193mm,top=12mm,paperheight=217mm]{geometry}

\begin{document}

\pagestyle{headings}
\mainmatter
\pagestyle{headings}
\mainmatter

\makeatletter
\newcommand{\printfnsymbol}[1]{%
  \textsuperscript{\@fnsymbol{#1}}%
}
\makeatother

\title{What Are Expected Queries in End-to-End Object Detection?} 

\titlerunning{What Are Expected Queries in End-to-End Object Detection?}
\authorrunning{S. Zhang, X. Wang, et al.}

\author{Shilong Zhang$^{1}$\thanks{Equal contribution. }, Xinjiang Wang$^{2}$\printfnsymbol{1}, \\ Jiaqi Wang$^{1}$, Jiangmiao Pang$^{1}$, Kai Chen$^{1,2}$}
\institute{$^1$Shanghai AI Laboratory
$^2$SenseTime Research \\\
\texttt{zhangshilong@pjlab.org.cn} 
\texttt{\{wjqdev,pangjiangmiao\}@gmail.com} \\
\texttt{\{wangxinjiang,chenkai\}@sensetime.com} \\
}

\maketitle

\vspace{-8mm}

\begin{abstract}

End-to-end object detection is rapidly progressed after the emergence of DETR. 
DETRs use a set of sparse queries that replace the dense candidate boxes in most traditional detectors.
In comparison, the sparse queries cannot guarantee a high recall as dense priors.  
However, making queries dense is not trivial in current frameworks. 
It not only suffers from heavy computational cost but also difficult optimization.
As both sparse and dense queries are imperfect, then \emph{what are expected queries in end-to-end object detection}?
This paper shows that the expected queries should be Dense Distinct Queries (DDQ).
Concretely, we introduce dense priors back to the framework to generate dense queries. 
A duplicate query removal pre-process is applied to these queries so that they are distinguishable from each other. The dense distinct queries are then iteratively processed to obtain final sparse outputs. We show that DDQ is stronger, more robust, and converges faster.
It obtains 44.5 AP on the MS COCO detection dataset with only 12 epochs. DDQ is also robust as it outperforms previous methods on both object detection and instance segmentation tasks on various datasets. DDQ blends advantages from traditional dense priors and recent end-to-end detectors. We hope it can serve as a new baseline and inspires researchers to revisit the complementarity between traditional methods and end-to-end detectors.  The source code is publicly available at \url{https://github.com/jshilong/DDQ}.

\keywords{Object detection, DETR, End-to-End, Query}
\end{abstract}

\section{Introduction}
\label{sec:intro}

\begin{figure}
\centering
\includegraphics[height=6.5cm]{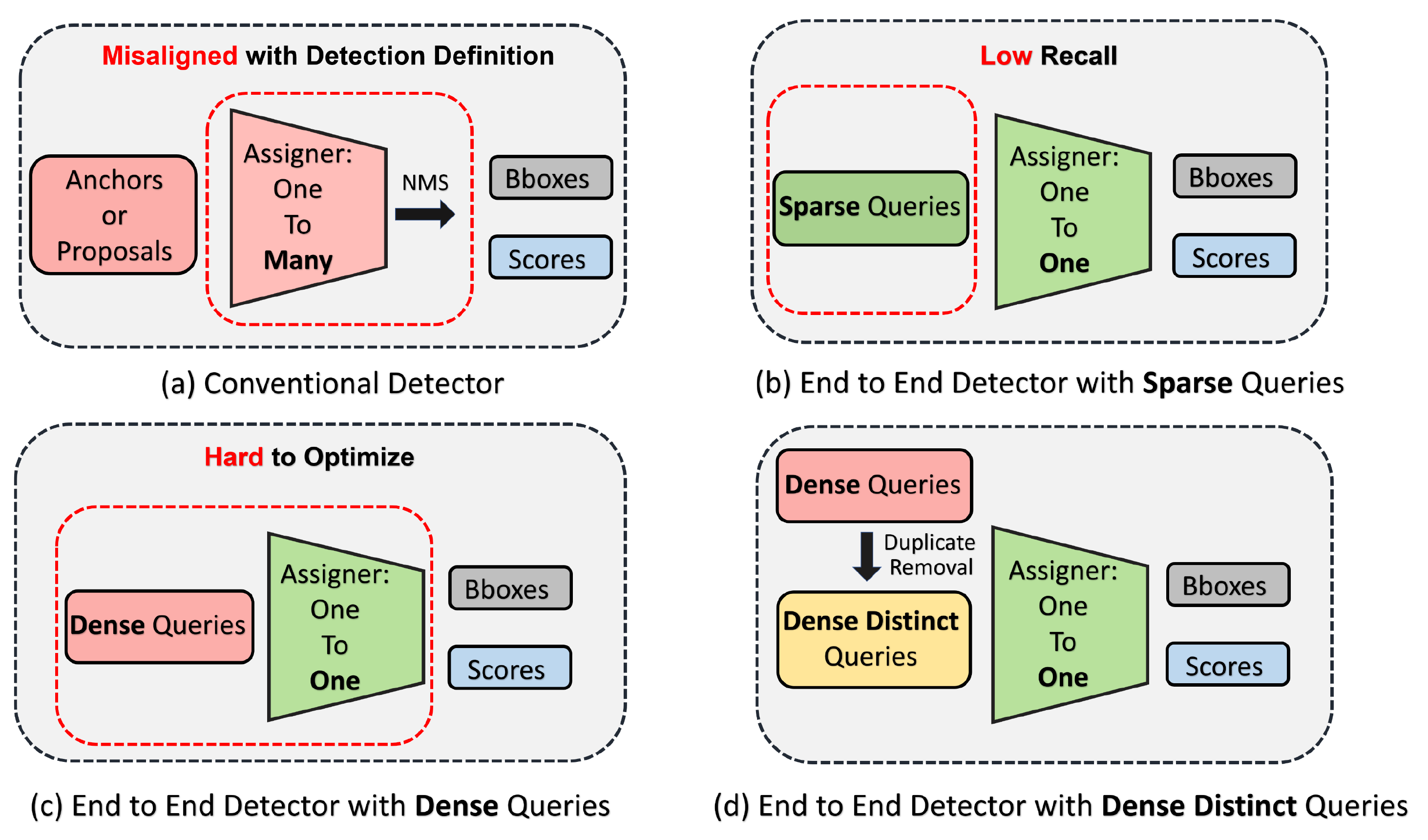}
\caption{(a) Dense priors in traditional object detectors no longer exist in (b) end-to-end object detectors. Intuitively dense queries can maintain a higher recall but are hard to implemented by (c) directly increasing the number of queries. (d) Alternatively, we use duplicate removal as a pre-process to obtain better dense distinct queries and keep the framework end-to-end.}
\label{fig:framework}
\vspace{-7mm}
\end{figure}

Object detection is one of the most fundamental challenges in computer vision which aims at localizing each object with a single bounding box. It brings a challenging problem that an accurate object detector should both detect all objects and avoid predicting duplicated boxes. 

To tackle this problem, the previous state-of-the-art methods~\cite{cai2018cascade,gao2021fast,ren2015faster} mostly follow a standard paradigm that first generates dense candidate boxes and assign one ground truth (GT) to many candidates for a high object recall, as shown in Fig.~\ref{fig:framework}(a). However, the one-to-many assignment results in redundant predictions. Since there should be only one prediction for each object in object detection, auxiliary post-processing, e.g., non-maximum suppression (NMS), is adopted to remove duplicated predictions. Although dominating object detection for years, this pipeline suffers from perfectly filtering out duplicated boxes without harming correct predictions.

This paradigm is broken by DETR~\cite{carion2020end}, an end-to-end object detection framework.
In contrast to the conventional paradigm, it throws away dense object candidates but directly initializes a set of sparse object queries.
When training, these queries are supervised by a one-to-one matching loss so that the optimization objective is consistent with the definition of object detection that only one bounding box is predicted for each object in the image. In this case, the network does not need post-processing to remove duplicated predictions anymore. 
However, DETR suffers from slow convergence speed which is explored in 
various works~\cite{gao2021fast,zhu2020deformable,sun2021sparse}.
One representative work following this paradigm is Sparse R-CNN~\cite{sun2021sparse}.
Sparse R-CNN interacts each query with a local region feature extracted by RoIAlign, leading to faster convergence speed compared to DETR.

Revisiting the framework of end-to-end object detectors that is sketched as Fig.~\ref{fig:framework}(b), there are only hundreds of sparse queries supervised by one-to-one matching loss. In this paper, we reveal that this design incurs a dilemma. On the one hand, the hundreds of sparse queries are not always sufficient to guarantee a high recall. On the other hand, when dense queries are introduced by directly increasing the number of queries to reach a higher recall, it will inevitably bring a lot of similar queries (as Fig.~\ref{fig:framework}(c)). These similar queries confuse the network since different labels are assigned to similar queries. This dilemma of choosing sparse or dense queries inspires us to think about \emph{what are expected queries in end-to-end object detection}?

This paper answers the question from a quantitative study and finally observes that the expected queries in end-to-end object detection should be \emph{dense distinct queries (DDQ)}, which means that the queries should be both densely distributed to detect all objects, as well as distinct from each other to facilitate the optimization of one-to-one matching loss. 

\begin{figure}[t]
\centering
\includegraphics[width=0.7\linewidth]{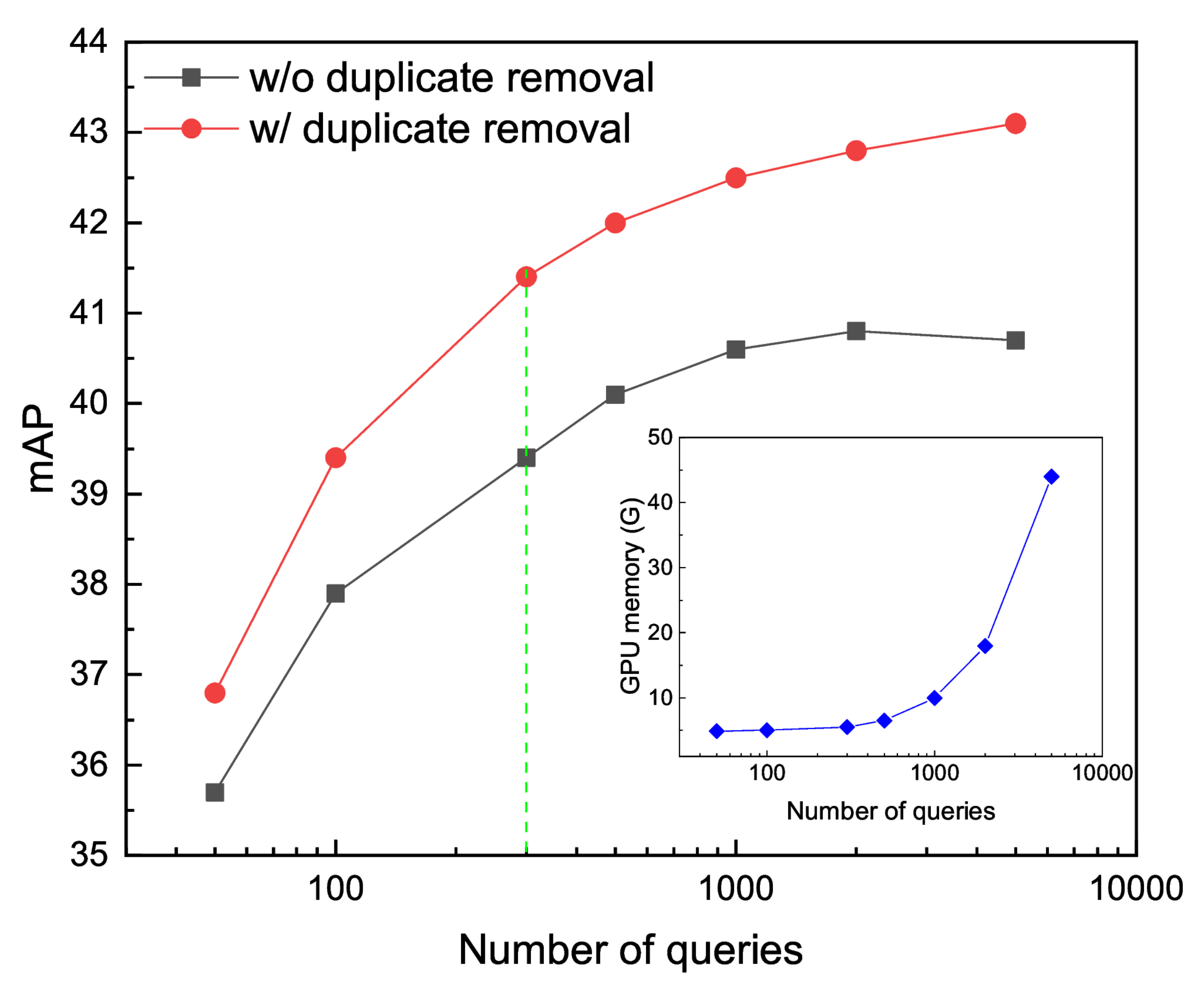}
\vspace{-3mm}
\caption{The performance comparison of Sparse R-CNN network with and without duplicate queries removal. All models are trained using the standard 1x setting. The green dotted line represents the default number(300) of queries adopted in Sparse R-CNN.  The subplot denotes the memory consumption per GPU as the number of queries increases in Sparse R-CNN. 
}
\label{fig:nms}
\vspace{-7mm}
\end{figure}

Specifically, as shown in Fig.~\ref{fig:nms}, we increase the number of queries in Sparse R-CNN. The performance increases at the beginning but finally plateaus and even decreases with denser queries because the training becomes more difficult with more similar queries.
By imposing a duplicate removal pre-processing to filter out similar queries and attain distinct queries before each stage of iterative refinements, the performance is improved with a clear margin.
More surprisingly, the performance margin consistently increases along with more queries\footnote{Similar trend is also observed in other end-to-end detectors such as Deformable DETR (see supplementary material).}{}.

Motivated by the performance of Sparse R-CNN with \emph{dense distinct queries} does not achieve a plateau with around seven thousand queries, we propose to introduce densely distributed queries on each location of the image which can be then converted to \emph{dense distinct queries}. These densely distributed queries guarantee a sufficient high recall to cover all potential target objects. 

However, directly processing densely distributed queries by the iterative refinement of Sparse R-CNN leads to unaffordable computational and GPU memory costs. 

As in Fig.~\ref{fig:nms}, when there are around seven thousand queries, Sparse R-CNN requires around 45G GPU memories, while there can be more than tens of thousand pixels on an image feature map.

Accordingly, based on Sparse-RCNN, we propose a novel framework, \emph{Dense Distinct Queries (DDQ)}, to introduce \emph{dense distinct queries} for end-to-end object detection and overcome the high computational cost.
Specifically, DDQ takes the feature point on each feature map as densely distributed initial queries. Instead of heavy RoI refinement heads, a lightweight fully convolutional network (ConvNet) is applied to process all queries in a sliding window manner, which shares a similar architecture as \cite{ren2015faster}. Differently, our design discards the anchor design and applies the bipartite matching algorithm to adaptively 
determine positive and negative samples for higher recall and robustness across different datasets. As a result, \emph{dense queries} are efficiently discriminated to generate \emph{dense distinct queries} with a reasonable computational cost. Moreover, a query distinctness enhancement mechanism further fuses these \emph{dense distinct queries} with their corresponding RoI features to enhance their distinctness. Different from Sparse R-CNN that requires 6 stages of iterative query refinement, DDQ achieves much higher performance with only 2 refinement stages and a fast convergence speed.

Experimental results evaluate the effectiveness and efficiency of the proposed method. DDQ pushes the frontier of state-of-the-art results on multiple object detection datasets. For example,  DDQ achieves 44.5 AP using a ResNet-50 backbone with a normal 1x training setting on MS-COCO~\cite{lin2014microsoft}, which largely surpasses the current state-of-the-art detectors (including both CNN-based and transformer-based) by over 2 AP with minimal inference time increase.  It also has leading performance on CrowdHuman~\cite{shao2018crowdhuman} with 93.2 AP and 98.2 Recall. For instance segmentation, DDQ also significantly outperforms Cascade Mask R-CNN~\cite{cai2018cascade} by 3 AP$_{mask}$ on MS COCO and 3.7 AP$_{mask}$ on LVIS v1.0 \cite{gupta2019lvis}.

\vspace{-3mm}

\section{Related Work}

\vspace{-1mm}

\label{sec:related_work}
\noindent\textbf{End-to-end Object Detection}
Designing an object detection framework free from post-processing operators such as NMS has been the target of many studies. Both DETR~\cite{carion2020end} and Sparse R-CNN~\cite{sun2021sparse} achieved the target by assigning a one-hot label for each ground truth in bipartite matching, and the cross attention module would then reason about the mutual relation to avoid duplicated predictions.  OneNet~\cite{sun2021rethinking} and DeFCN~\cite{wang2021end} are two representatives that developed NMS-free fully convolutional networks. Both of these works highlighted the importance of the one-to-one assignment during training for an object detector without duplicated predictions. However, all these networks need extra training time to learn a one-hot representation for each instance and are inferior in performance compared to traditional object detectors~\cite{cai2018cascade,feng2021tood,kim2020probabilistic}. These prior works justified the \emph{efficacy} of bipartite matching to learn to remove duplicated predictions in an end-to-end manner but ignored the \emph{efficiency} in achieving the goal. This study points out that similar queries are strong obstacles for bipartite matching to converge. Duplicated queries can be removed by class-agnostic NMS \emph{pre-processing} at each stage to relieve the burden of the bipartite matching.  This is also a different motivation from traditional detectors~\cite{cai2018cascade,girshick2015fast,ren2015faster} that adopt class-aware NMS as a \emph{post-processing} after the final prediction. 

\noindent\textbf{Query Design in DETR-like Structures} 
Object queries are a set of learned positional embeddings that guide the decoder to interact with the feature maps and positions to help infer the object and location at a specific region. It was first designed in ~\cite{carion2020end} to be a sparse and randomly initialized embedding set and learned through training, which was also adopted in the first stage in Sparse R-CNN~\cite{sun2021sparse}. Anchor DETR~\cite{wang2021anchor} provided correspondence between anchor points and query position. Conditional DETR ~\cite{meng2021conditional} introduced conditional spatial queries to help each cross-attention attend to the interesting regions inside a bounding box. DAB-DETR ~\cite{liu2021dab} explicitly learned a set of 4-D anchor boxes as queries. However, these studies are still based on sparse queries and would thus suffer from low recall problem, which is in contrast with this study. Noticeably, the two-stage version of Deformable DETR~\cite{zhu2020deformable} and Efficient DETR~\cite{yao2021efficient} also tried to introduce dense queries using the dense features from the last stage, which is similar to this study. However, both of these studies simply choose top-$K$ high-scoring queries out of the dense predictions and ignore the similarity among the kept queries. Therefore, the optimization difficulty problem due to similar queries is still not addressed. For example, the negligence of distinctness of queries made the performance quickly plateau as queries become denser and multiple refinement stages even hurt the performance as reported in Efficient DETR\cite{yao2021efficient}.

\vspace{-3mm}
\section{Dense Distinct Queries (DDQ)}
\vspace{-3mm}

\begin{figure}[t]
\centering
\includegraphics[width=12cm]{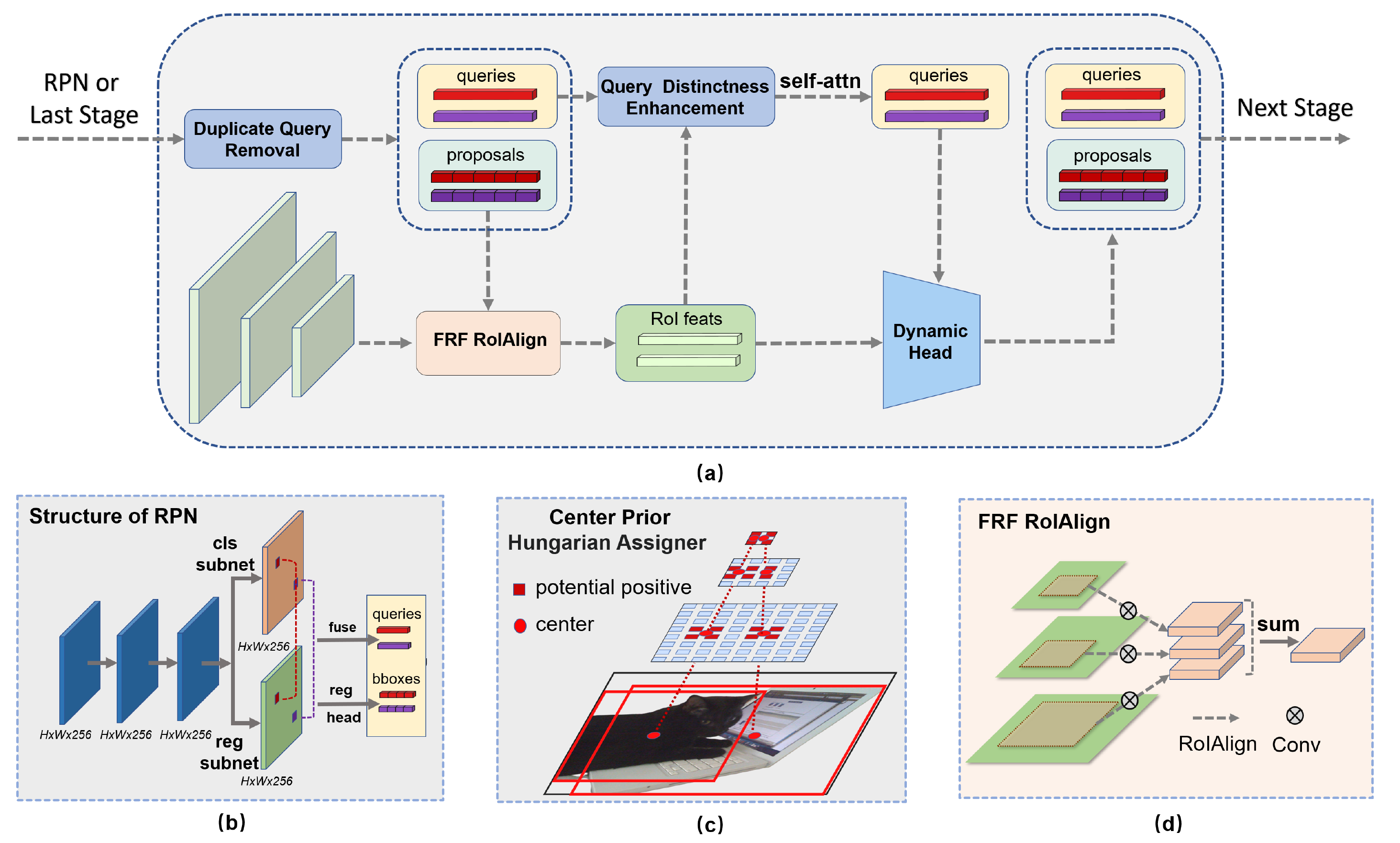}

\caption{The illustration of building blocks of our method. (a) The overall framework of DDQ: The input query first passes through a duplicate query removal process and is concatenated with the corresponding RoI feature to improve the distinctness of query. (b) The RPN framework: The classification and regression subnets share three sequential convolution layers and have one unique convolution layer to produce classification scores and regression offsets. (c): Assignment Strategy of RPN: The potential positive points are chosen with center prior. (d): Flexible Reception Field RoIAlign: RoIAlign features are pulled from neighboring feature levels to get flexible reception field for QFocal loss.}
\label{fig:pipeline}
\vspace{-6mm}
\end{figure}

\label{sec:method}
Dense distinct queries (DDQ) is the principle of designing an object detector with a fast convergence based on recent end-to-end detectors. Therefore, it is able to generalize to different architectures. The pipeline is sketched in Fig. \ref{fig:pipeline}. In the following sections, we describe the steps of applying dense distinct queries (DDQ) to Sparse R-CNN, one of the recent end-to-end object detectors with state-of-the-art performance.  

\vspace{-2mm}
\subsection{Revisiting Sparse R-CNN}

Sparse R-CNN mainly follows the paradigm of DETR and achieves better performance even without encoding layers thanks to its outstanding improvement in the decoding process. Sparse R-CNN utilizes dynamic instance interaction to replace the original cross-attention decoding part. Moreover, each object query in Sparse R-CNN only attends to features of a local region extracted by a RoIAlign operator instead of attending to all encoded features as in DETR. 

Sparse R-CNN maintains $N$ ($N\sim$300) independent queries with each corresponding to a bounding box. It then uses the bounding boxes to extract candidate region features through the RoIAlign operator from the feature pyramid.
Each query embedding is then used to generate convolutional parameters that interact with the RoI feature to output the predicted label and bounding box for each stage.

Sparse R-CNN also applies set prediction loss that utilizes bipartite matching according to the predefined matching cost to assign only one positive query for each ground truth. As discussed above, the sparse set of queries and duplicated queries are two bottlenecks for the performance and convergence of Sparse R-CNN.

\vspace{-4mm}
\subsection{Dense Queries}
It is described in Sec. \ref{sec:intro} that dense queries largely increase the recall rate while also bringing unacceptable computation cost. In this study,
a lightweight fully convolutional network (RPN) is applied to process all queries in a sliding window manner, The recall rate is largely increased with much smaller memory consumption thanks to the parameter sharing property of CNN structures. As traditional RPN used in \emph{e.g.} Faster R-CNN still lags in recall rate and also suffers from generalization issues due to its cumbersome anchor box design and assignment strategy, we present a new RPN structure to make it more efficient and robust. 

The RPN structure is shown as Fig. \ref{fig:pipeline}(b). 
Resembling single-stage detectors such as RetinaNet, the RPN structure in this study adopts $P_3$ to $P_7$ features, where $P_l$ represents the feature map level that is downsampled by $2^l$ from the input image size. It avoids the use of $P_2$ features as in the RPN structure in Faster R-CNN to save computation costs. It has 3 consecutive 3x3 Conv-GN-ReLU layers as a shared head structure, followed by two separate branches of one 3x3 Conv-GN-ReLU layer for classification and regression subtasks. The features from the two subnetworks are then extracted and concatenated to form dense queries such that each feature point is treated as a query. In this way, the number of queries  becomes much larger. For example, the number of queries reaches $13343$ given an image of size $800\times800$, which is two orders larger than that in Sparse R-CNN with only  minor increase in memory consumption.

We also discard the multiple anchor design and IoU-based assignment in the original RPN and apply the bipartite matching algorithm to adaptively discriminate positive and negative samples in order to increase the robustness ability across different datasets. Noticeably, the bipartite matching is slightly modified to only select positive samples out of the center feature points in a ground truth in order to stabilize training. 
Specifically, top-$K$ ($K=9$ in this study) nearest feature points to the center of ground truth on each level are regarded as potential positive samples, as shown in Fig.~\ref{fig:pipeline}(b). (More details about this RPN can be found in supplementary material.)

\vspace{-2mm}
\subsection{Distinct Queries}
We would first like to point out that the non-duplicate queries are of great importance to the convergence of the bipartite matching in end-to-end training methods such as Sparse R-CNN. As queries become similar, it is more difficult for the training to converge. 
This is understandable in the extreme case when there exist two identical queries. In this case, the bipartite matching assigns foreground label to one of them but background label to another. Without loss of generality, we adopt binary cross-entropy loss for classification. Therefore, the loss from these two queries becomes $L_1=-\log(p_1) - \log(1 - p_2)$, where $p_1$ and $p_2$ are the probability scores of the positive and negative query respectively, and satisfy $p_1 = p_2 = p$ as they are identical queries. In contrast, the loss value when only one of the duplicated queries exists is $L_0=-\log(p)$. The ratio of positive score gradient between the  duplicate and non-duplicate case is $\alpha$.
\begin{equation}
\vspace{-3mm}
\alpha = \frac{\partial L_1}{ \partial p} / \frac{\partial L_0}{ \partial p}  = 1 - \frac{p}{1-p} \
\vspace{1mm}
\end{equation}
It is obvious that gradient is scaled down ($\alpha < 1$) at $0<p<0.5$ and may even causes negative training ($\alpha < 0$) at $p > 0.5$.

\noindent
\textbf{Duplicate Query Removal} As shown in the toy example, the reduced gradient or even negative training caused by duplicated queries greatly suppresses the convergence. 
Therefore, we propose to remove duplicated queries as a pre-processing for each stage in Sparse R-CNN, as shown in Fig.~\ref{fig:pipeline}. 
Since each query represents a potential instance in an image, and an instance can be uniquely represented by its location in an image~\cite{wang2020solo}, it comes naturally to detect similar queries using the class-agnostic overlapping ratio of the corresponding bounding boxes. 
Therefore, the duplicate removal operator is realized by a \emph{class-agnostic} non-maximum suppression (NMS) in this study. 

It should be noted that the \emph{pre-processing} of queries is to relieve the burden of bipartite matching, which enables us to choose an aggressive  IoU threshold (defaults to 0.7 in this study, and performance Only fluctuates within 0.3 when it varies from 0.6 to 0.8) that is robust across different datasets. This pre-processing step maintains the advantages of end-to-end detectors that can align with detection definition. In contrast, traditional object detectors adopt class-aware NMS as a \emph{post-processing} after the final prediction, and the IoU threshold needs to be carefully tuned. 

\noindent
\textbf{Query Distinctness Enhancement} To make query features more discriminative, we enrich them with the extracted RoI features of the corresponding proposal box.
Each RoI feature is first average-pooled to the size of $1\times1$ and then concatenated with the original query, followed by a conv layer that reverts the channel number. As the RoI features contain more discriminative instance-level information compared to their corresponding queries, the combination further encourages the distinctness between different queries.
The enriched queries are then applied with self-attention to infer the mutual relations, followed by dynamic head modules to interact with the RoI features. This part follows the original design in Sparse R-CNN, and details are thus omitted in this study. 

\noindent
\textbf{Light-weighted Iterative Refinement}  Different from Sparse R-CNN that requires 6 stages of iterative query refinement, DDQ needs as few as 2 refinement stages.
Actually, the long iteration stages in Sparse R-CNN mainly compensate for the drawbacks caused by the independent sparse queries. On one hand, the corresponding region of initial sparse queries could be far from instances and thus needs long cascading stages to refine these queries, On the other hand, long refinement also helps distinguish similar queries to output one-hot prediction at each location. In contrast, the dense queries from RPN and duplicate removal pre-process before each stage addresses the above issues, and thus the number of iterative refinement can be significantly reduced without performance drop. 

\vspace{-3mm}

\subsection{Other Improvements}
\label{subsec:other_improvements}
\vspace{-2mm}

In this study, we also make extra efforts so that the network and optimization better align with \emph{dense distinct queries}.

\noindent
\textbf{Quality Focal Loss} We also follow recent one-stage methods~\cite{feng2021tood,li2021generalized} to adopt quality focal loss (QFL) that make IoU between bounding box predictions and gt bboxes as the target of classification. This modification is to better align the classification and regression subtasks for each query. The confidence score better reflects the regression quality and thus aids in the duplicate removal process using class-agnostic NMS. Other than the QFL classification, the regression loss function follows the design in Sparse R-CNN. 

\noindent
\textbf{RoIAlign with Flexible Receptive Field} Sparse R-CNN constrains each query to attend to only the RoIAligned region, which greatly decreases the computational overhead yet brings localized receptive field. Localized receptive field make model hard to to perceive the quality of bounding boxes. Hence, we design a efficient RoIAlign with Flexible Receptive Field (FRF), which combines extra RoIAligned features from neighboring levels in the feature pyramid, as shown in Fig.~\ref{fig:pipeline}(d). With the help of FRF RoIAlign, each query is attending to a wider range of features without introduction of heavy computations as \cite{guo2020augfpn}. FRF is also complementary to quality focal loss (QFL)  as the alignment of classification and regression in QFL need different scales of receptive fields to perceive the quality of bounding boxes.

\vspace{-5mm}
\section{Results}
\vspace{-4mm}
\label{sec:results}
In this section, we first show a progressive improvement in terms of both convergence speed and performance from Sparse R-CNN to DDQ. DDQ is also shown to significantly push the limit of object detection performance with various backbones, on different datasets and different tasks. 

\vspace{-5mm}
\subsection{Datasets}
\vspace{-2mm}
MS COCO 2017~\cite{lin2014microsoft} detection dataset is mainly used for comparison and ablation studies. It contains 118k training, 5k validation images, and 20k test images without annotations. There are on average 7 instances per image in this dataset. We report bounding box mean average precision (AP) as the performance metric, which is the average AP over multiple thresholds. If not specified, AP on the validation set is set as default. Apart from the object detection tasks, the instance segmentation performance on MS COCO is also reported in this study. 

LVIS v1.0~\cite{gupta2019lvis} is around the same size as MS COCO in regards to training images (100k) yet has a much larger number of classes (1203) and is deemed as a harder and long-tailed dataset to benchmark. Both object detection and instance segmentation results are benchmarked on LVIS v1.0 val with 20k images.

Besides, we also report the performance on CrowdHuman~\cite{shao2018crowdhuman} dataset, which has 15k training images and 4.4k validation images with around 23 heavily occluded instances per image, to demonstrate the robustness of DDQ on crowded scenes.

\vspace{-4mm}
\subsection{Implementation details}
\vspace{-2mm}
As for model training on MS COCO, ResNet-50~\cite{he2016deep} is the default backbone structure in this study if not specified. Most models adopt the 1x training protocol in MMDetection~\cite{chen2019mmdetection} with a mini-batch size 16 on 8 GPUS and a total training budget of 12 epochs. AdamW~\cite{loshchilov2018decoupled} optimizer with weight decay 0.05 is used, and the initial learning rate is $10^{-4}$ and decreases by 10 at epoch 9 and 11, respectively. It is worth noting that this setting does not bring improvement for the original Sparse R-CNN, and thus the comparison with Sparse R-CNN is deemed fair. All backbones are initialized with pretrained weights on ImageNet. Images are rescaled with short side 800 pixels, and only random horizontal flip are used for data augmentation. Loss weights are the same as previous work~\cite{carion2020end,sun2021sparse}. Some models are trained using the multi-scale training protocol with short side ranging from 480 to 800 pixels and elongated training budget of 24 epochs to compare with traditional CNN-based detectors. When comparing with DETR-based models, we also adopt the same augmentation in DETR in which random crop is involved and term this augmentation as \textit{DETR Aug}. We keep 300 queries from RPN and 200 queries after the first refinement stage, if not specified otherwise.

\vspace{-5mm}
\subsection{From Sparse R-CNN to DDQ}
\vspace{-3mm}
Table. \ref{tab:from_sparse_rcnn} shows a progressive development from Sparse R-CNN to DDQ in this study. Sparse R-CNN using 300 queries achieves 39.4 AP using the standard 1x training protocol, which is around 5.6 AP lower than that using 3x training time and heavier augmentations. The significant drop with short training time already implies the difficult convergence of Sparse R-CNN. Applying duplicate removal for queries at the beginning of each stage boosts the performance by 2 AP to 41.4 AP with little sacrifice in inference speed.  Further increasing the number of queries to 7000 increases the performance as well, yet with intimidating inference time. Replacing independent queries by the features generated by our developed RPN structure and cutting to 2 refinement stages maintains the performance of using 7000 queries but costs significantly less in both memory and inference time. Finally, DDQ is able to be comparable with Sparse R-CNN in latency but achieves 44.5 AP thanks to some other further structural improvements such as FRF RoIAlign and Query Distinctness Enhancement. This performance is ahead of the state-of-the-art object detectors by as much as 2 AP \cite{feng2021tood,dai2021dynamichead} that adopt the same backbone. The great improvement demonstrates the effectiveness of dense and distinct queries as a guiding principle of designing object detectors.

Notice that DDQ only increases marginal inference latency over Sparse R-CNN(17.7 ms vs 16.4 ms), which is much faster than other competing methods (Latency is benchmarked with Tesla A100, more detail can be found in supplementary material). For example Deformable DETR \cite{zhu2020deformable} achieves AP 43.8 AP with the latency 21.7 ms and Cascade R-CNN~\cite{cai2018cascade} achieves 40.3 AP with the latency 19.4 ms. DDQ both achieves better performance and inference faster than these methods.

\begin{table}[]
    \centering
    \vspace{-4mm}
    \caption{From Sparse R-CNN to DDQ: We show the results of adding Duplicate Query Removal(DQR), boosting the number of queries to 7000, and replacing the first 4 stages of Sparse R-CNN with our RPN, and finally, adding FRF RoI Align and QFocal to get DDQ. }
        \vspace{-2mm}
    \begin{tabular}{l|c|c|c|c|c|c|c}
    \hline
        Method & latency(ms) & AP & AP$_{50}$ & AP$_{75}$  & AP$_s$ & AP$_m$ &AP$_l$ \\
        \hline
        Sparse R-CNN  & 16.4 & 39.4 &57.7 & 42.5 & 22.4 & 41.8 & 54.3 \\
        \hline
        DQR & 16.8 & 41.4 & 60.9 &44.8 & 23.7 & 44.1 &56.3 \\
        \hline
        7000 Queries + DQR  & 114.9 & 43.0 & 62.7 & 46.8 & 27.5 & 46.0 & 57.1 \\
        \hline
        Change to RPN & 16.3 & 43.0 &62.2 & 47.4 & 26.6 & 45.6 &57.2 \\
        \hline
        DDQ & \textbf{17.7} & \textbf{44.5} & 63.3 & 48.8  & 27.8  & 47.3 & 58.0 \\

    \end{tabular}
    \vspace{-4mm}

    \label{tab:from_sparse_rcnn}
\end{table}
\vspace{-8mm}

\subsection{Comparison with Other Detectors}
Table. \ref{tab:comparison} shows the performance comparison with other latest object detectors using elongated training and heavier backbones. It is seen that DDQ with ResNet-50 backbone further increases from 44.5 AP to 47.7 AP when switching the training protocol from the standard 1x to 3x with DETR augmentations~\cite{carion2020end}. This performance largely surpasses all currently available end-to-end detectors by a clear margin while costing the smallest training budget. DDQ still remains its advantage among end-to-end object detectors as the backbone becomes heavier. As for the comparison with traditional CNN-based object detectors, we opt to the more commonly used multi-scale training with 24 epochs. We report the performance on COCO \textit{test} set for DDQ with ResNet-101 (R101) and ResNext-101-64d (X101-64d)~\cite{xie2017aggregated}. It is seen that the performance of DDQ also stays ahead even compared with the state-of-the-art CNN-based object detectors. It is also noted that DDQ can also benefit from additional encoder layers that are common in DETR-based detectors. For example, the performance of DDQ with R50 backbone and 3x DETR Aug reaches an amazing 49.8 AP when we add 6 dynamic blocks proposed in \cite{dai2021dynamichead} as the encoder, which is marked as \textbf{DDQ-R50-heavy} in Table.\ref{tab:comparison}.(For more details about DDQ-heavy see the supplementary material). 

\begin{table}[]
    \vspace{-6mm}
    \centering
        \caption{Comparison with other detectors including end-to-end detectors \emph{e.g.}, Conditional (Cond.) DETR~\cite{meng2021conditional}, Anchor DETR~\cite{wang2021anchor}, DAB DETR~\cite{liu2021dab}, Deformable (Deform.) DETR~\cite{zhu2020deformable}, Efficient DETR\cite{yao2021efficient}, Sparse R-CNN~\cite{sun2021sparse}, SMCA~\cite{gao2021fast} and traditional CNN-based object detectors \emph{e.g.}, Cascade R-CNN~\cite{cai2018cascade}, PAA~\cite{kim2020probabilistic}, TOOD~\cite{feng2021tood}, DyHead~\cite{dai2021dynamichead} on different training settings and with different backbones.}
   \vspace{-2mm}
    \begin{tabular}{l|c|c|c|c|c|c|c}
    \hline
        Method  & Epochs & AP & AP$_{50}$ & AP$_{75}$  & AP$_s$ & AP$_m$ & AP$_l$ \\
        \hline
        \hline
        \textit{DETR Aug\textbf{(Val)}}  &&&&&&& \\
        Cascade-R50 & 36 & 44.3 & 62.4 & 48 & 26.6 & 47.7 & 57.7\\
        Cond. DETR-R50 & 108 & 43.0 & 64.0 & 45.7 & 22.7 & 46.7 & 61.5 \\
        Anchor DETR-R50 & 50 & 42.1 & 63.1 & 44.9 & 22.3 & 46.2 & 60.0  \\
        DAB DETR-R50 & 50 &  42.6 & 63.2 & 45.6 & 21.8 & 46.2 & 61.1 \\ 
        Deform. DETR-R50 & 50  & 43.8 & 62.6 & 47.7 & 26.4 & 47.1 & 58.0 \\
        Efficient DETR-R50 & 36 & 44.2 & 62.2 & 48.0 & 28.4 & 47.5 & 56.6  \\

        Sparse R-CNN-R50 & 36 & 
        45.0 & 63.4 &48.2& 26.9& 47.2 &59.5 \\ 
        SMCA-R50 & 108 & 45.6 & 65.5 & 49.1 & 25.9 & 49.3 & 62.6 \\
        \textbf{DDQ-R50} & 36 & \textbf{47.7} &66.3 & 52.8 & 30.7 & 50.6 & 60.8 \\
        \textbf{DDQ-R50-heavy} & 36 & \textbf{49.8} &67.8 & 54.9 & 32.8 & 52.4 & 64 \\
        \hline
        Anchor DETR-R101 & 50 & 43.5 & 64.3 & 46.6 & 23.2 & 47.7 & 61.4  \\
        DAB DETR-R101 & 50 &  43.5 & 63.9 & 46.6 & 23.6 & 47.3 & 61.5 \\ 
        Sparse R-CNN-R101  &36 & 46.4 & 64.6 & 49.5   & 28.3  & 48.3 & 61.6 \\
        Efficient DETR-R101 & 36 & 45.2 & 63.7 & 48.8 & 28.8 & 49.1 & 59.0  \\
        
        \textbf{DDQ-R101} &36 & \textbf{48.5} & 67.1 & 53.2 & 32.2 & 51.5 & 61.8 \\
        \textbf{DDQ-R101-heavy} & 36 & \textbf{50.3} & 68.5 & 55.3  & 32.6 & 53.3 & 65.1 \\

        \hline
        \hline
        \textit{DETR Aug\textbf{(Test)}} &&&&&&& \\
        Sparse R-CNN-X101  & 36  & 46.9 & 66.3   &  51.2 & 28.6  & 49.2 & 58.7 \\
        Deform. DETR-X101  & 50  & 49 & 68.5  & 53.2 & 29.7 & 51.7 & 62.8  \\
        \textbf{DDQ-X101} & 36  & \textbf{50.0} & 68.8 & 55.0 & 32.1 & 52.4 & 61.4 \\
        \textbf{DDQ-X101-heavy} & 36  & \textbf{51.6} & 69.9 &  56.6 & 32.9 & 54.3 & 64.1 \\
        \hline
        \hline 
        \textit{Multi-scale Aug\textbf{(Test)}} &&&&&&& \\
        PAA-R101  & 24 & 44.8 & 63.3 & 48.7   & 26.5  & 48.8 & 56.3 \\
        TOOD-R101  & 24  & 46.7 & 64.6   & 50.7 & 28.9 & 49.6 & 57.0 \\
        DyHead-R101 & 24 & 46.5 & 64.5 & 50.7  & 28.3 & 50.3  & 57.5 \\
        \textbf{DDQ-R101 } &24 & \textbf{47.8} & 66.3 & 52.6 & 29.9 & 50.0 & 59.3 \\
        \hline 
        PAA-X101  & 24  & 46.6 & 65.6  & 50.8  & 28.8 & 50.4 & 57.9 \\
        TOOD-X101  & 24  & 48.3 & 66.5  & 52.4 & 30.7 & 51.3 & 58.6 \\
        DyHead-X101  & 24 & 47.7 & 65.7 & 51.9 & 31.5 & 51.7 & 60.7 \\
        \textbf{DDQ-X101} & 24  & \textbf{49.2} & 67.9 & 54.1 & 31.6 & 51.7 & 60.1 \\

    \end{tabular}

    \label{tab:comparison}
\end{table}

\vspace{-5mm}
\subsection{Results on The Other Datasets}
\vspace{-2mm}
To verify the robustness of DDQ on the other datasets, we report the performance comparison on CrowdHuman~\cite{shao2018crowdhuman} and LVIS v1.0~\cite{gupta2019lvis}. 
Following the standard-setting on CrowdHuman, the maximum possible detection of all  CNN methods is set to 500. For a fair comparison,  the number of kept queries is changed to 500 in DDQ and Sparse R-CNN~\cite{sun2021sparse}.The standard 1x setting on COCO is adopted when compared to CNN-based detectors. The setting with multi-scale input size of range 480-800 and total training epochs of 36, which is denoted as DDQ$^\dagger$, is also adopted for comparison with Sparse R-CNN$^\dagger$.
we use AP${50}$, mean miss rate (mMR) (the smaller the better), and recall rate as the evaluation metrics.   It is noted that DDQ leads in both settings and on every metric. 

A similar trend  is also observed on LVIS v1.0. And It is noteworthy that Sparse R-CNN gets poor performance on such a long-tailed  dataset. It lags significantly behind conventional CNN-detector. However, the performance of DDQ is still very stable and surpasses all methods by a large margin.

\begin{table}
\scriptsize
\begin{minipage}[c]{.48\linewidth}
\vspace{-0.4cm}
    \begin{center}
    \caption{Performance on CrowdHuman}
    \label{tab:realtime}
    \begin{tabular}{l|c|c|c|c}
    \hline
    Method &Epochs & AP$_{50}$& mMR & Recall \\
    \hline
    ATSS &12 & 87.0 &49.4 & 95.1 \\
    TOOD &12 & 88.7 & 46.5 &95.5 \\
    Cascade R-CNN &12 & 83.8 & 46.5& 97.5 \\
    DeFCN &32  & 89.1 & 48.9 & 96.5 \\
    \textbf{DDQ} &12 & \textbf{91.1} & \textbf{46.1} &\textbf{ 97.5} \\
    \hline
    Sparse R-CNN${^\dagger}$ &50 & 89.2 & 48.3 & 95.9 \\
    \textbf{DDQ$^\dagger$} &36  & \textbf{93.2} &\textbf{40.5 }& \textbf{98.2} \\
    \end{tabular}
    \end{center}
\vspace{-1cm}
\end{minipage}
\hspace{2mm}
\begin{minipage}[c]{.48\linewidth}
\vspace{-2.3cm}
    \caption{ Performance on LVIS v1.0 \textit{val} }

    \label{tab:deother}
    \vspace{-0.25cm}
    \begin{center}
    \begin{tabular}{l|c|c|c|c}
    \hline
        Method  & Epochs & AP & AP$_{50}$ & AP$_{75}$   \\
        \hline
        ATSS &12 & 22.6 & 32.7 & 23.9  \\
        Cascade R-CNN &12 & 24.6 & 36.1 & 26  \\
        Sparse R-CNN & 12 & 17.0 & 25.1 & 17.6  \\
        Sparse R-CNN & 36 & 21.4 & 30.2 & 22.5  \\
        DDQ & \textbf{12} & \textbf{28.2} & 39.5 & 30.3  \\
    \end{tabular}

    \end{center}
\vspace{-2.4cm}
\end{minipage}
\end{table}
We adopt the same mask head as QueryInst\cite{Fang_2021_ICCV} to compare  with other instance segmentation methods on MS-COCO \textit{test} dataset. As shown in Table~\ref{tab:instance_coco}, when trained using ResNet-50 backbone and same 680-800 multi-scale augmentations, DDQ outperforms QueryInst by 0.9 AP$_{mask}$ and 1.6 AP$_{bbox}$ with only $1/3$ training steps.  

QueryInst also cannot adapt to the long-tailed dataset LVIS v1.0, both detection and instance segmentation results lag significantly behind those of Cascade Mask R-CNN.
 DDQ achieves 29.6 AP$_{bbox}$ and 26.6 AP$_{mask}$, which are respectively 6.2 AP$_{bbox}$ and 5.2 AP$_{mask}$ higher than QueryInst and  4.1 AP$_{bbox}$ and 3.7 AP$_{mask}$ higher than Cascade Mask R-CNN~\cite{cai2018cascade} on LVIS v1.0.

\begin{table}
\scriptsize
\begin{minipage}[c]{.48\linewidth}
\vspace{-0.8cm}
    \begin{center}
    \caption{Instance segmentation Results on COCO \textit{test} dataset}
    \label{tab:instance_coco}
    \begin{tabular}{l|c|c|c}
    \hline
        Method  & Epochs & AP$_{bbox}$ & AP$_{mask}$   \\
        \hline
        Cascade Mask &36 & 44.5 & 38.6 \\

        QueryInst & 36 & 45.6 & 40.6   \\

        DDQ & \textbf{12} & \textbf{47.2} & \textbf{41.5}  \\

    \end{tabular}
    \end{center}
\vspace{-0.4cm}
\end{minipage}
\hspace{2mm}
\begin{minipage}[c]{.48\linewidth}
\vspace{-0.5cm}
    \caption{ Instance segmentation Results on LVIS v1.0 \textit{val} dataset}

    \label{tab:instance_lvis}
    \vspace{-0.25cm}
    \begin{center}
    \begin{tabular}{l|c|c|c}
    \hline
        Method  & Epochs & AP$_{bbox}$ & AP$_{mask}$ \\
        \hline
        Mask R-CNN  &12 & 22.5  & 21.7   \\
    
        Cascade Mask &12   & 25.5  & 22.9  \\
  
        QueryInst & 12 & 23.4 & 21.4  \\
     
        QueryInst & 36 & 22.5 & 20.8 \\
      
        DDQ & \textbf{12} & \textbf{29.6} & \textbf{26.6} \\

    \end{tabular}

    \end{center}
\vspace{-0.4cm}
\end{minipage}
\end{table}

\vspace{-5mm}
\section{Ablation study}
\vspace{-3mm}
\subsection{The Recall Improvement of Our RPN}
We analyze the recall with 0.5 IoU threshold, shown in Table \ref{tab:recall}. We regard the first 4 stages of Sparse R-CNN as RPN to make a fair comparison with our RPN. It can be found that the recall of vanilla RPN and using 300 queries in Sparse R-CNN  both have significant lower recall than that of our RPN design. Although having 7000 query can greatly improve the recall, the inference speed becomes extremely slow, which is 6.2 times slower than our method.

\begin{table}[]
    \centering
    \vspace{-0.18cm}
    \caption{Recall of RPN, DQR stand for Duplicate Query Removel }
    \begin{tabular}{l|c|c|c|c}

    \hline
        Method   & AR$_{100}$    & AR$_{200}$  & AR$_{300}$  & latency(ms)  \\
        \hline
        Sparse R-CNN  & 78.4  & 83.4  & 85.5  & 13.6 \\
        \hline
        7000 Queries \& DQR & 88.6 & 92.3  & 93.6  & \textbf{85.6} \\
        \hline
        Naive RPN & 76.4  & 83.0 & 86.2  & 19.2 \\
        \hline
        Our &  \textbf{87.6} & \textbf{91.1} & \textbf{92.7} & \textbf{13.7}  \\

    \end{tabular}
    \label{tab:recall}
\end{table}

\vspace{-8mm}
\subsection{Hyper-parameters in DDQ}
We find that the Center Prior Hungarian Assigner is not sensitive to the choice of   K. As shown in Table \ref{tab:k}, the performance fluctuations are all less than 0.3, only when we set k as a extremely large value such as 200, the training becomes very unstable. Therefore, we arbitrarily chose 9  and used this value in experiments on other datasets, and still achieved good results.

\begin{table}
\scriptsize
\begin{minipage}[c]{.48\linewidth}
    \vspace{-0.5cm}
    \begin{center}
    \caption{ Different K in Center Prior Hungarian Assigner}
    \label{tab:k}
    \vspace{-0.15cm}
    
    \begin{tabular}{l|c|c|c}
    \hline
        K   & AP  & AP$_{50}$ & AP$_{75}$ \\
        \hline
        1    & 44.1 & 62.7 & 48.3   \\

        5    & 44.4 & 62.9 & 48.7   \\
     
        9    & 44.5 & 63.3 & 48.8   \\
 
        13    & 44.6 & 63.2 & 48.5   \\

        100    & 44.2  & 63.0 & 48.5   \\
 
    \end{tabular}
    \end{center}
\vspace{-1.2cm}

\end{minipage}
\begin{minipage}[c]{.48\linewidth}
    \vspace{-0.7cm}
    \caption{Different number of convolutions in RPN}
    \label{tab:share_conv}
    \begin{center}
    \begin{tabular}{l|c|c|c}
    \hline
           & AP  & AP$_{50}$ & AP$_{75}$   \\
        \hline
        1    & 44.1 & 62.5 & 48.2  \\
 
        2    & 44.2 & 62.5 & 48.4   \\
    
        3    & 44.5 & 63.3 & 48.8 \\

        4    & 44.8 & 63.3  & 49.3  \\
 
        5    & 44.6 & 63.3 & 49.2   \\

    \end{tabular}
    \end{center}
\vspace{-1.4cm}
\end{minipage}
\end{table}

\begin{table}
\scriptsize
\begin{minipage}[c]{.48\linewidth}
\vspace{-0.5cm}
    \begin{center}
     \caption{Performance with different number of stages(S) and queries(Q).}
    \label{tab:queryandstage}
    
    \begin{tabular}{l|c|c|c|c|c}
    \hline
          & 100 Q   & 200 Q & 300 Q & 400 Q & 500 Q \\
      \hline
        S=1  &  43.1 & 43.3  & 43.4 & 43.5 & 43.5 \\
      
        S=2  &  43.6 & 44.1  & 44.5 & 44.6 & 44.5 \\
   
        S=3  &  43.9 & 44.7  & 45.0 & 45.1 & 44.9 \\
 
        S=4  &  43.9 & 44.6  & 44.9 & 45.1 & 45.0 \\

    \end{tabular}
    \end{center}
\vspace{-0.4cm}
\end{minipage}
\hspace{2mm}
\begin{minipage}[c]{.48\linewidth}
\vspace{-0.5cm}
    \caption{ Impact of how to construct query in RPN.}

    \label{tab:constrctquery}
    \begin{center}

    \begin{tabular}{l|c|c|c}
    \hline
            &AP  & AP$_{50}$ & AP$_{75}$   \\
        \hline
        None   &44.1 &62.6 &48.5  \\
     
        FPN    & 44.2 & 62.5 & 48.4   \\
     
        share conv  & 44.4 & 62.8 &48.9  \\
      
        cls\&reg   & 44.5 & 63.3 & 48.8 \\

    \end{tabular}
    \end{center}
\vspace{-0.4cm}
\end{minipage}
\end{table}

\vspace{-5mm}
The RPN is quite robust for the number of share convolutions.  As shown in Table \ref{tab:share_conv}, it can be more lightweight when you decrease the number to 2 and get higher performance with 4 convolutions.

We analyze the combination of different number of stages and different number of queries. Table ~\ref{tab:queryandstage} shows that the best number of stages is proportional to the number of queries. This is easy to understand. When the number of queries increases, although the recall increases, the quality of the newly added queries is low, and more stages are needed to improve the quality of queries. It worth emphasizing that we use 2 stages and 300 queries to balance the performance and latency and when using 3 stages with the same number queries, our method achieves even higher performance 45.0 AP on MS COCO.

We try four ways to construct the query in our RPN. As shown in Table \ref{tab:constrctquery}, None means all queries are set to a zero tensor and the refinement stage only get meaningful query bounding boxes. This attempt drops the performance to 44.1 AP. Simply constructing queries from the FPN results in 0.3 AP degrade. Constructing queries from the last share convolution can be an alternative as it can get a comparable performance.

\subsection{Influence of Other Components in DDQ}

As shown in Table \ref{table:ablation_qe}, the Query Distinctness Enhancement(QDE) increases the performance by 0.7 AP as a standalone plugin. However, QFcoal loss and FRF Align are observed to only function well when combined, and either one alone only brings slight improvement. This is consistent with the analysis in Sec. \ref{subsec:other_improvements} that FRF RoIAlign is able to help QFocal loss distinguish the quality of queries through the fusion of multi-layer RoI features. Therefore, FRF RoIAlign has a synergy effect with QFocal loss.  

\begin{table}[]
    \centering

    \caption{Impact of Query Distinctness Enhancement(QDE), Flexible Reception Field RoIAlign(FRF)  and QFocal loss(QFL).}

    \label{table:ablation_qe}
    \begin{tabular}{ccc|cccccc}
    \hline
        \textbf{QDE} & \textbf{FRF} &  \textbf{QFL} & AP & AP$_{50}$ & AP$_{75}$ & AP$_s$ & AP$_m$ & AP$_l$ \\  \hline
                    &             &     &    43.0  &    62.1   & 47.4   &26.6   &45.6 & 56.3  \\ \hline
        \checkmark  &             &     &   \textbf{43.7}  &    62.8   &48.0    &26.7   &46.2 & 57.2    \\ 
                    &  \checkmark &     &   43.2  &    61.9   &47.2    &27.3   &46.0 & 55.6     \\ 
                    &             &\checkmark &  43.2   &   62.3 &47.9  &26.1   &46.1 &57.0    \\ \hline
                    &  \checkmark &\checkmark &   \textbf{44.0} &62.6 &48.5 &27.3 &47.0 &57.8    \\
        \checkmark  & \checkmark  &     &    44.0  &62.9 &48.1 &27.1 &47.1 &57.0    \\
                    
         \checkmark &             &\checkmark &   43.9 & 63.0 &48.4 & 26.9 & 46.9 &57.0   \\ \hline
         \checkmark &  \checkmark &\checkmark &   \textbf{44.5} &63.1 &48.9 &27.2 &47.6 &57.7    \\
    \vspace{-4mm}
    \end{tabular}
\end{table}

\vspace{-12mm}
\section{Conclusions}
\vspace{-3mm}

This paper answer the question ``what are the expected queries for end-to-end object detection". Based on  quantitative analysis on Sparse R-CNN, we conclude that the expected queries should be simultaneously dense and distinct. The entire framework, dubbed as DDQ, is stronger, converges faster, and is more robust across different datasets. 
DDQ blends advantages from the traditional dense priors and the recent end-to-end detectors. 
We hope it can serve as a new baseline and inspires researchers to think about the complementarity between traditional methods and end-to-end detectors.

\clearpage

\bibliographystyle{splncs04}
\bibliography{egbib}

\begin{thebibliography}{10}
\providecommand{\url}[1]{\texttt{#1}}
\providecommand{\urlprefix}{URL }
\providecommand{\doi}[1]{https://doi.org/#1}

\bibitem{cai2018cascade}
Cai, Z., Vasconcelos, N.: Cascade r-cnn: Delving into high quality object
  detection. In: Proceedings of the IEEE conference on computer vision and
  pattern recognition. pp. 6154--6162 (2018)

\bibitem{carion2020end}
Carion, N., Massa, F., Synnaeve, G., Usunier, N., Kirillov, A., Zagoruyko, S.:
  End-to-end object detection with transformers. In: European Conference on
  Computer Vision. pp. 213--229. Springer (2020)

\bibitem{chen2019mmdetection}
Chen, K., Wang, J., Pang, J., Cao, Y., Xiong, Y., Li, X., Sun, S., Feng, W.,
  Liu, Z., Xu, J., et~al.: Mmdetection: Open mmlab detection toolbox and
  benchmark. arXiv preprint arXiv:1906.07155  (2019)

\bibitem{dai2021dynamichead}
Dai, X., Chen, Y., Xiao, B., Chen, D., Liu, M., Yuan, L., Zhang, L.: Dynamic
  head: Unifying object detection heads with attentions. In: Proceedings of the
  IEEE/CVF Conference on Computer Vision and Pattern Recognition. pp.
  7373--7382 (2021)

\bibitem{Fang_2021_ICCV}
Fang, Y., Yang, S., Wang, X., Li, Y., Fang, C., Shan, Y., Feng, B., Liu, W.:
  Instances as queries. In: Proceedings of the IEEE/CVF International
  Conference on Computer Vision (ICCV). pp. 6910--6919 (October 2021)

\bibitem{feng2021tood}
Feng, C., Zhong, Y., Gao, Y., Scott, M.R., Huang, W.: Tood: Task-aligned
  one-stage object detection. In: Proceedings of the IEEE/CVF International
  Conference on Computer Vision. pp. 3510--3519 (2021)

\bibitem{gao2021fast}
Gao, P., Zheng, M., Wang, X., Dai, J., Li, H.: Fast convergence of detr with
  spatially modulated co-attention. arXiv preprint arXiv:2101.07448  (2021)

\bibitem{girshick2015fast}
Girshick, R.: Fast r-cnn. In: Proceedings of the IEEE international conference
  on computer vision. pp. 1440--1448 (2015)

\bibitem{guo2020augfpn}
Guo, C., Fan, B., Zhang, Q., Xiang, S., Pan, C.: Augfpn: Improving multi-scale
  feature learning for object detection. In: Proceedings of the IEEE/CVF
  Conference on Computer Vision and Pattern Recognition. pp. 12595--12604
  (2020)

\bibitem{gupta2019lvis}
Gupta, A., Dollar, P., Girshick, R.: Lvis: A dataset for large vocabulary
  instance segmentation. In: Proceedings of the IEEE/CVF conference on computer
  vision and pattern recognition. pp. 5356--5364 (2019)

\bibitem{he2016deep}
He, K., Zhang, X., Ren, S., Sun, J.: Deep residual learning for image
  recognition. In: Proceedings of the IEEE conference on computer vision and
  pattern recognition. pp. 770--778 (2016)

\bibitem{kim2020probabilistic}
Kim, K., Lee, H.S.: Probabilistic anchor assignment with iou prediction for
  object detection. In: European Conference on Computer Vision. pp. 355--371.
  Springer (2020)

\bibitem{li2021generalized}
Li, X., Wang, W., Hu, X., Li, J., Tang, J., Yang, J.: Generalized focal loss
  v2: Learning reliable localization quality estimation for dense object
  detection. In: Proceedings of the IEEE/CVF Conference on Computer Vision and
  Pattern Recognition. pp. 11632--11641 (2021)

\bibitem{lin2014microsoft}
Lin, T.Y., Maire, M., Belongie, S., Hays, J., Perona, P., Ramanan, D.,
  Doll{\'a}r, P., Zitnick, C.L.: Microsoft coco: Common objects in context. In:
  European conference on computer vision. pp. 740--755. Springer (2014)

\bibitem{liu2021dab}
Liu, S., Li, F., Zhang, H., Yang, X., Qi, X., Su, H., Zhu, J., Zhang, L.:
  Dab-detr: Dynamic anchor boxes are better queries for detr. In: International
  Conference on Learning Representations (2021)

\bibitem{loshchilov2018decoupled}
Loshchilov, I., Hutter, F.: Decoupled weight decay regularization. In:
  International Conference on Learning Representations (2018)

\bibitem{meng2021conditional}
Meng, D., Chen, X., Fan, Z., Zeng, G., Li, H., Yuan, Y., Sun, L., Wang, J.:
  Conditional detr for fast training convergence. In: Proceedings of the
  IEEE/CVF International Conference on Computer Vision. pp. 3651--3660 (2021)

\bibitem{ren2015faster}
Ren, S., He, K., Girshick, R., Sun, J.: Faster r-cnn: Towards real-time object
  detection with region proposal networks. Advances in neural information
  processing systems  \textbf{28},  91--99 (2015)

\bibitem{shao2018crowdhuman}
Shao, S., Zhao, Z., Li, B., Xiao, T., Yu, G., Zhang, X., Sun, J.: Crowdhuman: A
  benchmark for detecting human in a crowd. arXiv preprint arXiv:1805.00123
  (2018)

\bibitem{sun2021sparse}
Sun, P., Zhang, R., Jiang, Y., Kong, T., Xu, C., Zhan, W., Tomizuka, M., Li,
  L., Yuan, Z., Wang, C., et~al.: Sparse r-cnn: End-to-end object detection
  with learnable proposals. In: Proceedings of the IEEE/CVF Conference on
  Computer Vision and Pattern Recognition. pp. 14454--14463 (2021)

\bibitem{sun2021rethinking}
Sun, Z., Cao, S., Yang, Y., Kitani, K.M.: Rethinking transformer-based set
  prediction for object detection. In: Proceedings of the IEEE/CVF
  International Conference on Computer Vision. pp. 3611--3620 (2021)

\bibitem{wang2021end}
Wang, J., Song, L., Li, Z., Sun, H., Sun, J., Zheng, N.: End-to-end object
  detection with fully convolutional network. In: Proceedings of the IEEE/CVF
  Conference on Computer Vision and Pattern Recognition. pp. 15849--15858
  (2021)

\bibitem{wang2020solo}
Wang, X., Kong, T., Shen, C., Jiang, Y., Li, L.: Solo: Segmenting objects by
  locations. In: European Conference on Computer Vision. pp. 649--665. Springer
  (2020)

\bibitem{wang2021anchor}
Wang, Y., Zhang, X., Yang, T., Sun, J.: Anchor detr: Query design for
  transformer-based detector. arXiv preprint arXiv:2109.07107  (2021)

\bibitem{xie2017aggregated}
Xie, S., Girshick, R., Doll{\'a}r, P., Tu, Z., He, K.: Aggregated residual
  transformations for deep neural networks. In: Proceedings of the IEEE
  conference on computer vision and pattern recognition. pp. 1492--1500 (2017)

\bibitem{yao2021efficient}
Yao, Z., Ai, J., Li, B., Zhang, C.: Efficient detr: improving end-to-end object
  detector with dense prior. arXiv preprint arXiv:2104.01318  (2021)

\bibitem{zhu2020deformable}
Zhu, X., Su, W., Lu, L., Li, B., Wang, X., Dai, J.: Deformable detr: Deformable
  transformers for end-to-end object detection. arXiv preprint arXiv:2010.04159
   (2020)

\end{thebibliography}
\end{document}